\documentclass{vgtc}

\ifpdf
  \pdfoutput=1\relax                   
  \usepackage{graphicx}                
  \DeclareGraphicsExtensions{.pdf,.png,.jpg,.jpeg} 
\else
  \usepackage{graphicx}                
  \DeclareGraphicsExtensions{.eps}     
\fi%

\graphicspath{{figures/}{pictures/}{images/}{./}} 

\usepackage{microtype}                 
\PassOptionsToPackage{warn}{textcomp}  
\usepackage{textcomp}                  
\usepackage{mathptmx}                  
\usepackage{times}                     
\usepackage{cite}                      
\usepackage{tabu}                      
\usepackage{booktabs}                  
\usepackage{xcolor}
\usepackage{amsmath}
\usepackage{wrapfig}
\usepackage{xspace}
\usepackage{enumitem}
\usepackage[numbers]{natbib}
\usepackage[normalem]{ulem}
\usepackage{titlesec}

\onlineid{0}

\definecolor{sysred}{HTML}{E47475}
\definecolor{sysblue}{HTML}{2979FF}
\definecolor{editgreen}{HTML}{3B8031}

\newcommand{\feature}{\textit{Feature Distribution View}}
\newcommand{\strip}{\textit{Subgroup Overview}}
\newcommand{\detail}{\textit{Detailed Comparison View}}
\newcommand{\suggest}{\textit{Suggested and Similar Subgroup View}}
\newcommand{\system}{\textsc{FairVis}\xspace}

\titlespacing\section{0pt}{8pt minus 2pt}{3pt plus 1pt minus 2pt}

\vgtccategory{Research}
\vgtcinsertpkg

\title{\system{}: Visual Analytics for\\Discovering Intersectional Bias in Machine Learning}

\author{\'Angel Alexander Cabrera
\and Will Epperson
\and Fred Hohman
\and Minsuk Kahng
\hspace{10mm}
\and Jamie Morgenstern
\and Duen Horng (Polo) Chau\thanks{E-mails: \{acabrera30, willepp, fredhohman, kahng, jamiemmt.cs, polo\}@gatech.edu}}
\affiliation{\scriptsize Georgia Institute of Technology}

\abstract{
The growing capability and accessibility of machine learning has led to its application to many real-world domains and data about people. 
Despite the benefits algorithmic systems may bring, models can reflect, inject, or exacerbate implicit and explicit societal biases into their outputs, disadvantaging certain demographic subgroups.
Discovering which biases a machine learning model has introduced is a great challenge, due to the numerous definitions of fairness and the large number of potentially impacted subgroups.
We present \system{}, a mixed-initiative visual analytics system that integrates a novel subgroup discovery technique for users to audit the fairness of machine learning models.
Through \system{}, users can apply domain knowledge to generate and investigate known subgroups, and explore suggested and similar subgroups.
\system{}'s coordinated views enable users to explore a high-level overview of subgroup performance and subsequently drill down into detailed investigation of
specific subgroups.
We show how \system{} helps to discover biases in two real datasets used in predicting income and recidivism.
As a visual analytics system devoted to discovering bias in machine learning, \system{} demonstrates how interactive visualization may help data scientists and the general public understand and create more equitable algorithmic systems.
} 

\keywords{Machine learning fairness, visual analytics, intersectional bias, subgroup discovery}

\teaser{
  \centering
  \includegraphics[width=.95\linewidth]{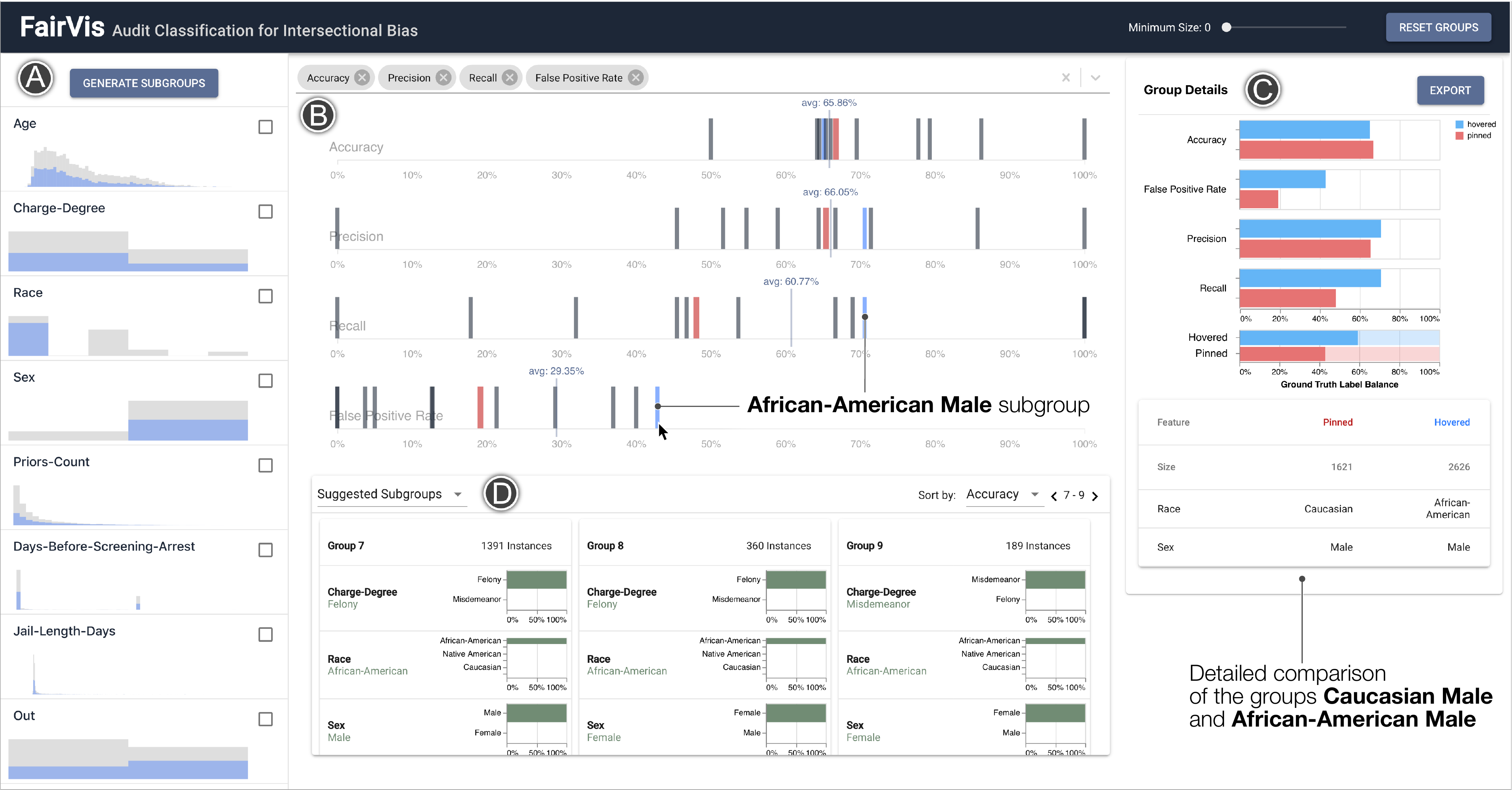}
  \vspace{-9pt}
  \caption{
    \system{} integrates multiple coordinated views for discovering intersectional bias. 
    Above, our user investigates the intersectional subgroups of \textit{sex} and \textit{race}. 
    \textbf{A.}
    The \feature{} allows users to visualize each feature's distribution and generate subgroups.
    \textbf{B.}
    The \strip{} lets users select various fairness metrics to see the global average per metric
    and compare subgroups to one another, e.g.,
    \textcolor{sysred}{pinned \textbf{Caucasian Males}} versus \textcolor{sysblue}{hovered \textbf{African-American Males}}.
    The plots for \textit{Recall} and \textit{False Positive Rate} show that for African-American Males, the model has relatively high recall but also the highest false positive rate out of all subgroups of sex and race. 
    \textbf{C.} 
    The \detail{} lets users compare the details of two groups and investigate their class balances.
    Since the difference in False Positive Rates between Caucasian Males and African-American Males is far larger than their difference in base rates, a user suspects this part of the model merits further inquiry. 
    \textbf{D.} 
    The~\suggest{} shows suggested subgroups ranked by the worst performance in a given metric. 
  }
	\label{fig:teaser}
}

\begin{document}


\firstsection{Introduction}

\maketitle

In recent years, significant strides have been made in machine learning (ML), enabling automated, data-driven systems to tackle ever more challenging and complex tasks. 
Many of the new domains in which these novel techniques are being applied are human-focused and consequential, including hiring, predictive policing, predicting criminal recidivism, and pedestrian detection.
The latter two cases are examples where differing levels of predictive accuracy have been observed for different demographic groups~\cite{wilson2019predictive,chouldechova2017fair}.

When deploying ML to these societally impactful domains,
it is vital to understand how models are performing on all different
types of people and populations. ML algorithms are usually
trained to maximize the overall accuracy and performance of their
model, but often do not take into account disparities in performance
between populations. The trained models thus provide no guarantees as
to how well they will perform on different subgroups of a dataset.

The potential disparity in performance between populations may have
many sources; an ML model can naturally encode implicit
and explicit societal biases~\cite{barocas2016big}, which is often
referred to as algorithmic bias. Performance disparity can arise for a
variety of reasons: the training data may not be representative,
either in terms of its representation of different demographic groups
or within a particular demographic group; the training data labels may
have errors which reflect societal biases, or be an imperfect proxy
for the ultimate learning task; unequal rates of labels across
demographic groups; the model class may be overly simple to capture
more nuanced relationships between features for certain groups; and
more~\cite{chouldechova2017fair}.  A stark example of algorithmic bias in
deployed systems was discovered by Buolamwini and Gebru's Gender
Shades study~\cite{buolamwini2018gender}, who showed that many commercially
available gender classification systems from facial image data had accuracy
gaps of over 30\% between darker skinned women and lighter skinned
men.  While the overall models' accuracies hovered around 90\%, darker
skinned women were classified with accuracy as low as 65\% while the
models' accuracies on lighter skinned men were nearly 100\%.

\begin{figure}
    \centering
    \includegraphics[width=0.69 \linewidth]{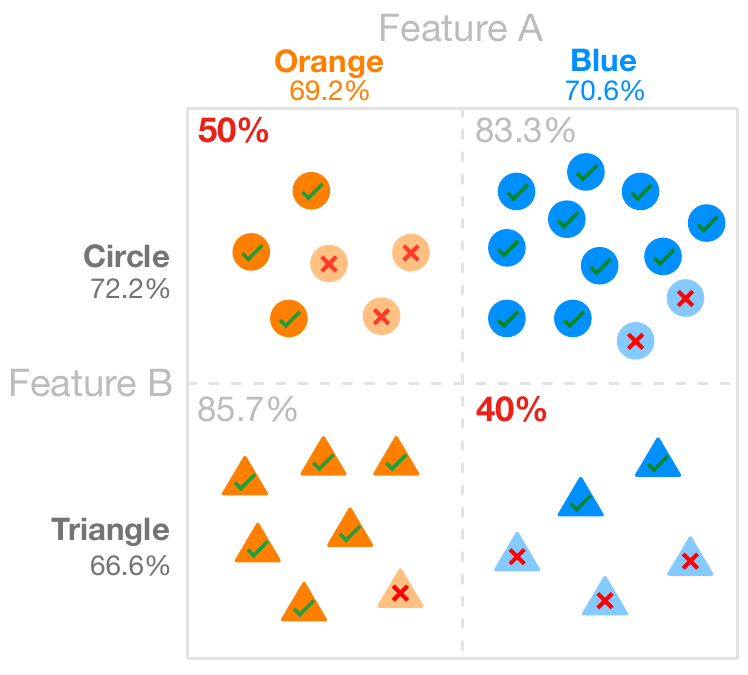}
    \vspace{-14pt}
    \caption{This illustrative example highlights how inequities in populations can be masked by aggregate metrics. 
    While the classifier in this example has an accuracy of between 66.6\% and 72.2\% when looking at groups defined by a single feature, the accuracy drops to as low as 40\% when looking at the intersectional subgroups.}
    \label{fig:intersectionality}
\end{figure}

In order to discover and address potential issues before ML systems
are deployed, it is vital to audit ML models for algorithmic bias.
Unfortunately, discovering biases can be a daunting task, often due to
the inherent intersectionality of bias as shown
by Buolamwini and Gebru~\cite{buolamwini2018gender}. Intersectional bias is bias that is
present when looking at populations that are defined by multiple
features, for example ``Black Females'' instead of just people who are
``Black'' or ``Female''. The difficulty in finding intersectional bias
is pronounced in the Gender Shades study introduced above --- while
there were performance differences when looking at sex and skin color
individually, the significant gaps in performance were only found when
looking at the intersection of the two features. An example of how
aggregated measures can hide intersectional bias can be seen
in~\autoref{fig:intersectionality}.

In addition to the intersectional nature of bias, addressing bias is
challenging due to the numerous proposed definitions of
unfairness.
The metrics for measuring a model's fairness
include measuring a model's group-specific false positive rates,
calibration, and more. 
While a user may decide on one or
more metrics to focus on, achieving true algorithmic fairness can
be an insurmountable challenge. In \autoref{background}, we describe
how recent research has shown that it is often impossible to fulfill
multiple definitions of fairness at once.

While it can be straightforward to audit for intersectional bias when
looking at a small number of features and a single fairness
definition, it becomes much more challenging with a large number of
potential groups and multiple metrics. When investigating
intersectional bias of more than a few features, the number of
populations grows combinatorially and quickly becomes
unmanageable. Data scientists often have to balance the tradeoffs between various fairness metrics when making changes to their models.

To help data scientists better audit their models for intersectional bias, we introduce \textbf{\system{}}, 
a novel visual analytics system dedicated to helping audit the fairness of ML models.
\system{}'s major contributions include:

\begin{itemize} [topsep=1mm, itemsep=0mm, parsep=2mm, leftmargin=4mm]
\item \textbf{Visual analytics system for discovering intersectional
    bias.}  
    \system{} is a mixed-initiative system that
  allows users to explore both suggested and user-specified
  subgroups that incorporate a user's existing domain knowledge.  
  Users can visualize how these groups rank on various common
  fairness and performance metrics and contextualize subgroup
  performance in terms of other groups and overall performance.  Additionally, users can compare the feature distributions
  of groups to make hypotheses about why their performance
  differs.  Lastly, users can explore similar subgroups to compare
  metrics and feature values.
    
\item \textbf{Novel subgroup generation technique.}  In order to aid
  users in exploring a combinatorially large number of subgroups, we
  introduce a new subgroup generation technique to recommend
  intersectional groups on which a model may be underperforming.  We
  first run clustering on the training dataset to find statistically
  similar subgroups of instances.  Next, we use an entropy technique
  to find important features and calculate fairness metrics for the
  clusters.  Lastly, we present users with the generated subgroups
  sorted by important and anomalously low fairness metrics.  These
  automated suggestions can aid users in discovering subgroups on
  which a model is underperforming.
    
\item \textbf{Method for similar subgroup discovery.}  Once a
  subgroup for which a model has poor performance has been
  identified, it can be useful to look at similar subgroups to compare
  their values and performance.  We use similarity in the form of
  statistical divergence between feature distributions to find
  subgroups that are statistically similar.  
  Users can then compare
  similar groups to discover which value differences impact
  performance or to form more general subgroups of fewer features.
\end{itemize}

\section{Background in Machine Learning Fairness}\label{background}
Significant discoveries and advances have been made in algorithmic
bias detection, mitigation, and machine learning fairness in recent years.  Most of the work
stems from theoretical computer scientists and sociologists
focusing on the mathematical foundations and societal impacts of
machine learning.

A major difficulty in machine learning fairness is that it is
mathematically impossible to fulfill all definitions of fairness
simultaneously when populations have different base rates.  This
incompatibility between fairness metrics was formalized by the
\textit{impossibility theorem} for fair machine learning.  Two papers
~\cite{kleinberg2017inherent, friedler2016possibility} simultaneously
proved that if groups have different base rates in their labels, it is
statistically impossible to ensure fairness across three base fairness
metrics --- balance for the positive class, balance for the negative
class, and calibration of the model.  Data scientists must therefore
decide which fairness metrics to prioritize in a model and how to make
trade-offs between metric performance.

The implications of this discovery were made apparent in the
recidivism prediction tool COMPAS, a system that is used to predict
the risk of letting someone go on bail.  A ProPublica
article~\cite{angwin2016bias} showed that COMPAS is more likely to
rank a Black defendant as higher risk than a White defendant given
that they have equal base rates.  A follow-up study showed that while
COMPAS is not balanced for the positive class prediction, it is well
calibrated, meaning that the model provides similarly accurate scores
for both groups relative to their base
rates~\cite{dieterich2016compas}.  Due to inherent base rate
differences, it is not possible for COMPAS to meet the all three
fairness definitions at once. We explore this dataset more in \autoref{case-1}.

There have been various solutions proposed for addressing algorithmic
bias in machine learning across the entire model training pipeline.
These range from techniques for obfuscating sensitive variables in
training data~\cite{xu2018fairgan}, to new regularization parameters
for training~\cite{beutel2019putting} and post-processing outcomes by
adding noise to predictions~\cite{hardt2016equality}.  While these can
help balance certain inequities, the impossibility theorem dictates
that hard decisions will still have to be made about which fairness
metrics are the most important for each problem.  Ideally, over time
these will become standard processes for ensuring model fairness, and
tools like \system{} can be used to ensure their effectiveness and
investigate tradeoffs between metrics.

\section{Related Work}
\subsection{Intersectional Bias}

Important innovations have come from the machine learning community in relation to intersectional bias.

Kearns et al.~\cite{kearns2018preventing} proposes a framework for auditing a
(possibly very large) number of subgroups for unfair treatment. Their
work has the same high-level concerns that motivate this project: that
there may be a very large number of intersectional groups over which
one wants to satisfy some notion of fairness. However, for their work,
they assume the collection of these groups is predefined for the task
at hand, and construct an algorithm for creating a distribution over
classifiers which (approximately) minimizes a particular fairness
metric over all the subgroups simultaneously. Our work differs from
theirs in several key ways. First, we aim to operate in a space where
a predefined notion of groups is not necessarily available, and so
cooperation between an automated system and a domain expert might be
necessary to uncover subgroups whose treatment by a particular model
is problematic. 
Second, our goal is to help a user explore their model and dataset for a deeper understanding of \emph{why} the model might be treating particular groups very differently, a far different task compared to aiming to 
satisfy a particular fairness metric without delving into the data-dependent sources of this different treatment.
This deeper model understanding will facilitate task-specific interventions and promote a deeper understanding of a learning task, a dataset's suitability to this task, and whether a model (class) matches the dataset and task.

Recent techniques have also been proposed for discovering and
analyzing intersectional bias.  Most similar to our work is Slice
Finder~\cite{chung18slice}, a technique for automatically generating
subgroups.  Slice Finder takes a top-down approach to generating
subgroups, adding features to create more granular groups until the
training loss is statistically significant.  Our technique for
automated subgroup discovery is bottom-up, clustering instances
without imposing any structure on the features used.  In addition to
potentially generating more diverse subgroups, our subgroups are not
tied to training loss, allowing us to use any performance metric to
order and suggest subgroups.

\subsection{Visual Analytics for Machine Learning}

There is a large body of work on visual analytics techniques
for understanding and developing machine learning models~\cite{amershi2015modeltracker, kahng2016visual,
  krause2017workflow, patel2010gestalt, ren2017squares}. Various
systems have been created focused on helping users understand how
complex models work and visually debugging their
outputs~\cite{seq2seqvisv1}.  Additionally, systems have been
introduced that enable users to analyze production-level
models~\cite{kahng2018activis} and the full ML workflow from training to
production~\cite{krause2017workflow, patel2010gestalt,
  amershi2015modeltracker}.  These visual systems have been shown to
aid users in understanding and developing machine learning
models~\cite{hohman2018visual}.

The research most directly related to the present work are techniques
for analyzing both datasets and the results of machine learning
models.  MLCube~\cite{kahng2016visual} is a visualization technique
that allows users to compare the accuracy of groups defined by at most
two features.  Squares~\cite{ren2017squares} introduces a novel
encoding for visually understanding the performance of a multi-class
classifier.  Finally, Facets~\cite{facets} is a visualization system
for interactively exploring and subdividing large datasets.  While
these systems provide novel and useful methods for exploring data and
outcomes, they are limited by the complexity of subgroups and number
of performance metrics they support, essential features for auditing
for intersectional bias.

While there are various visual systems for analyzing machine learning
models, there have been few advancements in visual systems or
techniques focused on algorithmic bias and fair machine learning.  One
notable exception is the \textit{What-If} tool from Google~\cite{whatif}.  The
\textit{What-If} tool is a more general data exploration tool that combines
dataset exploration with counterfactual explanations and fairness
modifications.  Users can explore a dataset using the Facets
interface, and then look for
counterfactual~\cite{kusner2017counterfactual} explanations for
specific instances.  There is also a feature that allows users to modify a
classifier's threshold to change which fairness principles are being
satisfied.  While the \textit{What-If} tool is a powerful data exploration
tool, it does not allow users to explore intersectional bias nor does it aid
users in auditing the performance of specific subgroups.

\section{Design Challenges and Goals} 

Our goal is to build an interactive visual interface to help users explore the fairness of their machine learning models and discover potential biases.
Many of the challenges present in auditing for bias derive from the combinatorial number of subgroups generated when looking at various features.
Additionally, any visual system must convey multiple fairness metrics for a subgroup.
A successful visual system should allow users to
narrow the large search space of possible subgroups.
We formalize these important factors in the design of \system{} with the following key design challenges:

\subsection{Design Challenges} \label{challenges}

\textbf{C1. Auditing the performance of known subgroups.} For many datasets and problem definitions, users already know of certain populations for which they want to ensure fair outcomes. \cite{vealeFairness}  It is often cumbersome and slow to manually generate and calculate various performance metrics for subgroups. A system should enable users to generate any type of subgroup they want to investigate, and efficiently generate and calculate metrics for it \cite{holstein2019improving}.
    
\textbf{C2. Contextualizing subgroup performance in relation to multiple metrics and other groups.} 
To measure the severity of  bias against a certain subgroup, it is important to know how the subgroup is performing in relation to the overall model. 
Any visual encoding of subgroup performance should convey how groups perform for different performance metrics \cite{hardt2016equality} and in relation to other subgroups. 
Our interface should also allow users to drill down into subgroup details while maintaining the high-level view.
    
\textbf{C3. Discovering significant subgroups in a large search space.} 
When investigating intersectional bias, there could be hundreds or thousands of subgroups a user may need to look at \cite{kearns2018preventing}.
It is often not feasible to analyze every group, so deciding how to prioritize subgroups is an important and difficult task.
Methods for discovering and suggesting potential groups can aid users in searching this large space and finding potential issues more efficiently.
    
\textbf{C4. Finding similar subgroups to investigate feature importance and more general groups.}
When a biased subgroup has been identified, it can be informative to look at the performance of similar subgroups to draw conclusions about feature importance or to create more general groups \cite{dworkSimilarFairness, zemelSimilarFairness}. 
This is a difficult task since an immense number of potential subgroups have to be searched to find similar subgroups, and it is not clear how similarity between subgroups should be defined or calculated.

\textbf{C5. Emphasizing the inherent trade-offs between fairness metrics.} Classifiers are often not able to fulfill all measures of fairness if the base rates between populations are different, as proven by the \textit{impossibility of fairness} theorem (\autoref{background}).
This means users often have to keep in mind the tradeoffs between fairness metrics when deciding what modifications to make to their models.
It is essential to show the various fairness metrics when displaying subgroup performance and emphasize their tradeoffs.

\textbf{C6. Suggesting potential causes of biased behavior.} How to address bias in machine learning models is a difficult and open question, but there are indicators that can help users start to improve their models.
Emphasizing information like ground-truth label balance, subgroup entropy, and data distribution can point users in the right direction for addressing biases \cite{hardt2016equality, krauseUserStudy}.

\subsection{Design Goals}
Using the design challenges we identified for ML bias discovery, we iterated and developed design goals for \system{}.
The following goals address the challenges presented in \autoref{challenges}, and align with the primary interface components of our system: 

\textbf{G1. Fast generation of user-specified subgroups.}
Since users often have domain knowledge about important subgroups they want to ensure fairness for (C1), quickly generating these groups to enable investigation is vital. Users should be able to select either entire features (e.g. ``race'') or specific values (e.g. ``white'' or ``black'') to generate groups of any feature combination (C3). 
Users should then be able to explore the performance of these groups in detail.

\textbf{G2. Combined overview relationships with detailed information of subgroup performance.}
To understand the magnitude and type of bias a model has encoded for a subgroup, it is important to show the performance of the group in relation to the overall and other subgroups' performance (C2).
At the same time, the interface should also display detailed information about the performance of the selected subgroup (C6). 
We aim to achieve this by using multiple, coordinated views that can handle different fairness metrics (C5).

\textbf{G3. Suggested under-performing subgroups for user investigation.} When more than a couple of features are used to define subgroups, the number of generated groups grows combinatorially
(C3). We aim to develop both an algorithmic technique for automatically discovering potentially under-performing subgroups and an intuitive visual encoding for suggesting discovered groups to the user. By suggesting these groups automatically, we can make the subgroup discovery process quicker and potentially discover groups the user had not originally thought about (C2).

\textbf{G4. Efficient calculation of similar subgroups.}
For any given subgroup, there is a combinatorially large space of groups that need to be searched to find similar groups (C3). Since it is often useful to look at similar subgroups to analyze the importance of certain features or to generate more general groups, we aim to develop a technique that efficiently discover these similar groups (C4, C6).

\textbf{G5. Effective visual interfaces for subgroup comparison.}
Users may want to analyze two subgroups side by side to compare their values or performance (C2). We aim to provide an intuitive  interface for highlighting the differences between two groups. Users can compare these groups to help pinpoint which features or values are causing the difference in fairness metrics (C6).

\section{\system{}: Discovering Intersectional Bias}

From the design challenges introduced in Section 4, we have developed \system{}, a visual analytics system for discovering intersectional bias in machine learning models. 
To meet the listed design goals, we developed two novel techniques to generate underperforming subgroups and find similar subgroups. 
We combine these techniques in a web-based system that tightly integrates multiple, coordinated views to help users discover fairness issues in known and unknown subgroups.

Our interface consists of four primary views, the \feature~(\autoref{sec:feature}), \strip~(\autoref{sec:overview}), \suggest~(\autoref{sec:suggest}, \autoref{sec:similar}), and \detail~(\autoref{sec:detail}). The \feature~gives users an overview of the dataset distribution and allows them to generate groups to visualize in the \strip. Users can then add additional subgroups provided by the \suggest, and compare and further analyze them in the~\detail. Each section of our interface aligns with one of the stated design goals, addressing each desired feature.

\subsection{Feature Distribution View \& Subgroup Creation [G1]} 
\label{sec:feature}

The left sidebar, or~\feature, acts as both a high-level overview of a dataset's distribution and the interface for generating user-specified subgroups.
As a starting place for \system{}, the~\feature~helps users develop an idea of their dataset's makeup and begin auditing subgroups right away.

\begin{figure}
    \centering
    \includegraphics[width=.87\linewidth]{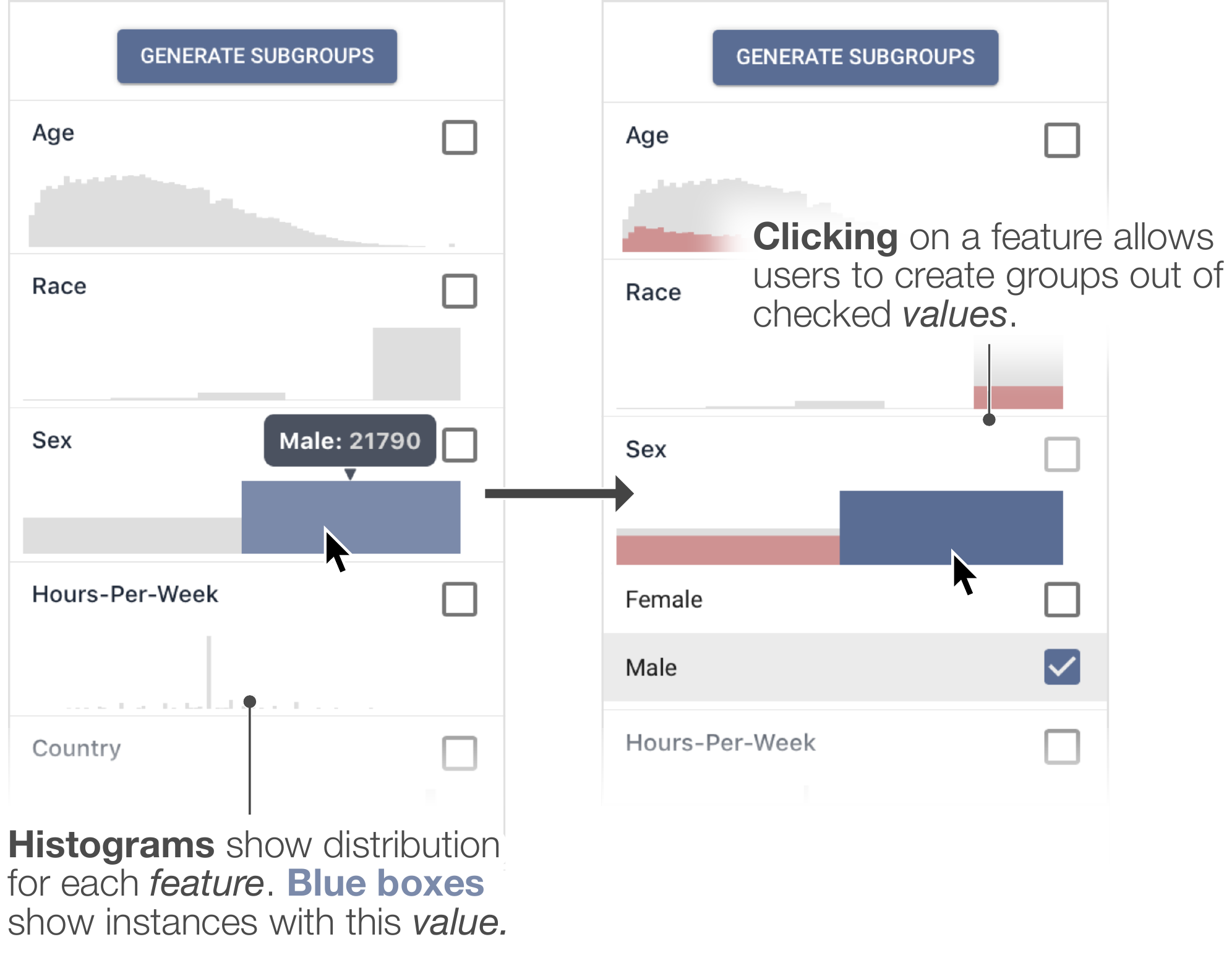}
    \vspace{-10pt}
    \caption{The~\feature~allows users to explore both the distributions of each feature in the entire dataset and also create user-specified groups out of features or specific values. When a user hovers over a bar such as ``Male'', it shows the number of instances for that value. Red bars show the distribution of the pinned group (in this case ``White Males'') from the~\strip~. }
    \label{fig:drawer}
\end{figure}
\textbf{Feature distribution.} A large part of understanding model performance is understanding how the data used to train a model is distributed (C6). 
We enable users to investigate feature distributions by providing large, interactive histograms for each feature for the entire dataset, as seen in \autoref{fig:drawer}.
These histograms treat all features as categorical and when a user hovers over a bar, a tooltip shows the value of this category and how many instances there are with that value in the entire dataset.
Furthermore, clicking on one of the rows reveals a collapsible view of all the possible values for the feature.
Users are also able to hover over the expanded values to see their location in the histogram.
\begin{figure*}
    \centering
    \includegraphics[width=\textwidth,height=5.4cm,keepaspectratio]{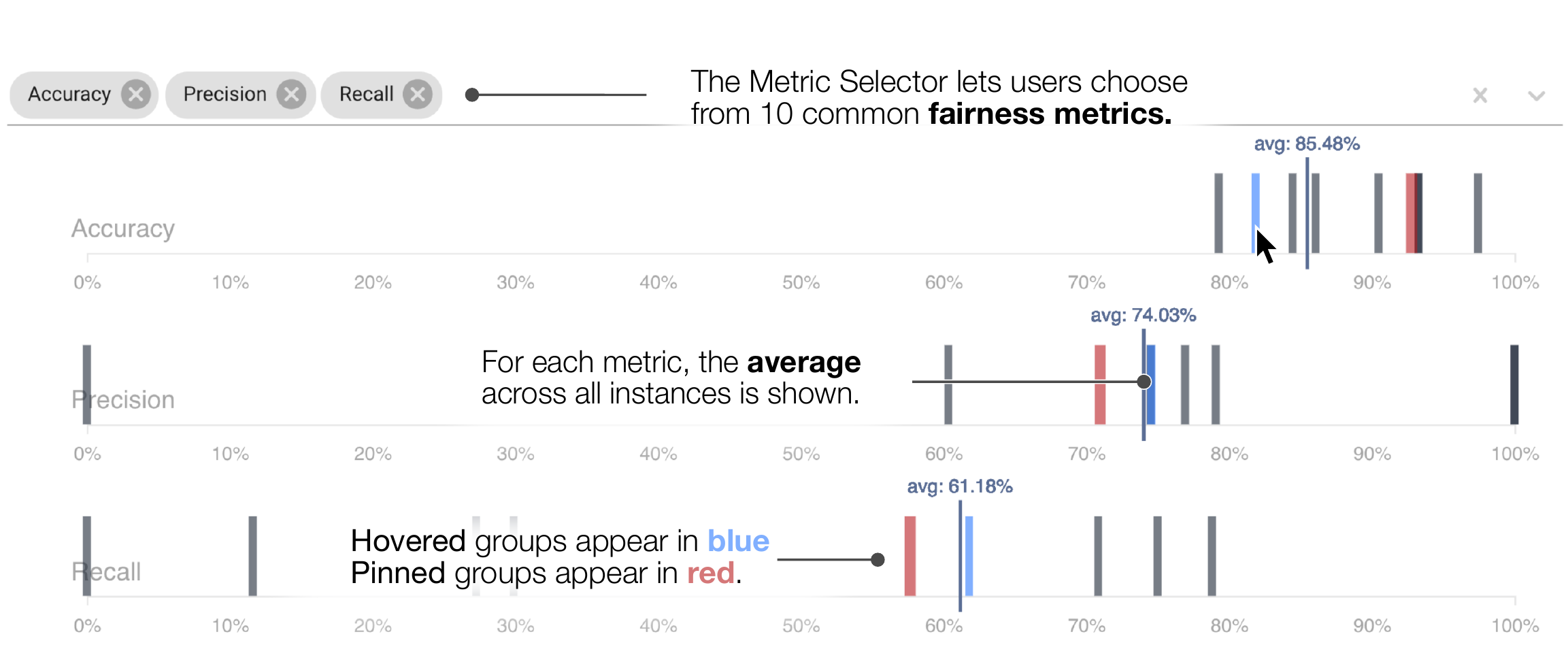}
    \vspace{-14pt}
    \caption{In the~\strip~users can see how different subgroups compare to one another according to various performance metrics. As more metrics are selected at the top, additional strip plots are added to the interface. Here, a user has \textbf{\textcolor{sysred}{pinned}} the Female subgroup and \textbf{\textcolor{sysblue}{hovers}} over the Male subgroup.}
    \label{fig:stripfig}
\end{figure*}

\textbf{Subgroup generation.} The~\feature~also allows users to generate user-specified subgroups. 
Model developers are often aware of certain intersectional subgroups for which they want to ensure fairness (C1). 
We define a subgroup as a subset of a dataset in which all instances share certain values, e.g., the subgroup of blue circles in~\autoref{fig:intersectionality}.

Our interface allows users to generate both specific subgroups and all subgroups of multiple features by selecting a combination of features and values.
For instance in \autoref{fig:drawer}, if a user checks the feature ``race'' and ``sex'', then mutually exclusive subgroups will be generated out of all the instances in the dataset divided on their values for ``race'' and ``sex''.
However, if a user wants to investigate a particular subgroup, they can select a specific value for ``race'' and ``sex'' to add a subgroup of all instances with those specific values.
Users can pick any number and combination of features and values by which to define their subgroups, and thus are at liberty to define how general or specific the subgroups they want to explore are.

\subsection{Subgroup Overview [G2]} \label{sec:overview}

Once a user has generated subgroups, they should be able to understand which subgroups the model is underperforming on across various metrics and further investigate interesting subgroups (C2).
The~\strip~provides a high-level view of this information as multiple interactive and dynamic strip plots (C2).

When a user clicks the ``Generate Subgroups'' button (\autoref{fig:drawer}), \system{} splits the data into the specified subgroups and calculates various performance metrics for them.
These groups are then represented in the multiple strip plots as lines corresponding to their performance for the respective metric.

\textbf{Visualizing multiple fairness metrics.} Due to the inherent tradeoffs between different fairness requirements as shown by the \textit{impossibility theorem}, users must choose which metrics they want to prioritize and investigate (C5).
To facilitate this interaction, we allow users to select which metrics are displayed in the~\strip~by adding and removing performance metrics through the bar seen at the top of ~\autoref{fig:stripfig}.
Selecting a new metric adds an additional strip plot for that metric with all the current subgroups.
We also show the corresponding dataset average per metric in each strip plot to provide context as to how each subgroup is doing in relation to the overall dataset.

In total, users can select from the following metrics: Accuracy, Recall, Specificity, Precision, Negative Predictive Value, False Negative Rate, False Positive Rate, False Discovery Rate, False Omission Rate, and F1 score.
These metrics were selected as they are typically the most common metrics used for evaluating the equity and performance of classification models. 
The performance metrics are derived from the same base outcome rates of true positives, true negatives, false positives, and false negatives.
If users find that they need different metrics for performance, they can add a new definition using the base rates which are available in the system.

When a user hovers over a subgroup in a strip plot, the corresponding group is highlighted on every plot currently displayed. 
This allows users to see how an individual group performs on several different metrics at once [C2, C5].
To further investigate a subgroup, the user can click on a bar to pin the group and use the~\detail~to further investigate the group.

\textbf{Choice of visual encoding.} 
We chose a strip plot to visualize performance metrics since it allows users to focus on the relative magnitude of subgroup performance in relation to other subgroups and the overall dataset performance.
By juxtaposing plots, users are able to see how different metrics are spread out \cite{juxtapositionVisual}.
One of the shortcomings of strip plots is that they can become crowded and hard to use with a large number of subgroups.
We address this issue by allowing users to filter the strip plot by subgroup size. 
While subgroups come in all sizes, groups that are only a few instances are usually not statistically significant enough to draw conclusions from.
The size filtering mechanism can help users narrow their search space (C3) and improve the functionality of the strip plot.

While designing our system we considered different visual encodings for displaying subgroups, especially a scatterplot matrix.
We decided to use a strip plot over a scatterplot matrix for several reasons. 
First, since each of the performance metrics is derived from the same base rates, many of the relationships between metrics are arithmetic and not indicative of interesting patterns. We investigated outliers and found that they did not systematically represent any interesting subgroups.
Additionally, scatterplot matrices redundantly encode information, as every metric is displayed multiple times.
Our strip plot implementation only includes each metric once while still allowing users to see how the group performs in regards to other metrics.
Multiple strip plots allow us to display the most important information in a clean and understandable manner; namely, how a given subgroup is performing for selected metrics and in relation to the overall dataset and other subgroups.

\subsection{Suggested Subgroups [G3]} \label{sec:suggest}

While many users may know of certain groups in their dataset they need to ensure fairness for, it is possible that the model developer has little domain knowledge and does not know where to start.
Since there are a combinatorially large number of subgroups in a dataset, it is daunting and often times not feasible to manually inspect groups for every combination of features.

To help the user find potentially biased subgroups, we generate
subgroups algorithmically and present them to the user for
investigation.  
The~\suggest~at the bottom of the interface displays
these subgroups and allows the user to sort them by any fairness
metric to discover underperforming subgroups (C3).

\subsubsection{Generating and Describing Suggested Subgroups}

To create the suggested subgroups, we use a clustering-based generation technique. By clustering instances, we can generate groups with significant statistical similarity that can be described by a few dominant features.
We can subsequently calculate their performance metrics and display them to the user.

We first cluster all the data instances by their feature values in one-hot encoded form.
We use K-means as our clustering algorithm~\cite{hartigan1979algorithm} with K-means++ as the seeding~\cite{arthur2007k}.
Users are able to choose the hyperparameter $K$ to balance the number and size of generated subgroups --- a smaller $K$ produces larger, less defined groups while a larger $K$ has the opposite effect. 
Users run the clustering as a pre-processing script before uploading their data to \system{}.

We also experimented with more sophisticated clustering algorithms like the density-based algorithms DBSCAN and OPTICS, which can generate arbitrarily shaped and sized clusters. 
While the statistical quality of the density-based clusters can be higher, we found that the flexibility provided by allowing users to modify $K$ is more helpful for discovering important and useful subgroups. Additionally, we found that since we were clustering on many one hot encoded categorical features, DBSCAN's notion of density was not as useful and K-means produced higher quality clusters. Given prior successful application of K-means to a variety of problems and tasks with both categorical and numerical features, we decided to first adapt K-means for \system{} 
\cite{oyelade2010application, de2008clustering}.

Once the clusters have been generated, the makeup of the group must be described to the user.
A cluster's instances are made up of a variety of values for each feature, but some features may be more dominated by one value than others.
We define a \textit{dominated feature} as a feature that consists of mostly one value, the \textit{dominant value} in a subgroup.
For example, if a cluster is 99\% male for the feature sex, sex is a dominated feature with a dominant value of male.

The most dominant features can be used to describe the makeup of a subgroup to the users.
We rank how dominant features of a group are by calculating the entropy of each feature distribution over its values.
Entropy is used since it describes how uniform a feature is.
The closer a feature's entropy is to 0, the more concentrated the feature is in one value, making it more dominant in that subgroup.

We formalize the technique for finding dominant features as follows. Suppose we have a set of features, $\mathcal{F} = \{ f_1, f_2, ..., f_i, ...\}$, with
each feature, $f_i$, having a set of possible values, $V_i = \{ v_{i1}, v_{i2}, ...\}$.
We calculate the \textit{feature entropy} for
the $k$-th subgroup and $i$-th feature, $S_{k, i}$, as follows: 
\vspace{-7pt}
\begin{align}
    S_{k,i} = - \sum_{v\in V_i} \frac{N_{k,v}}{N_k} \log{\frac{N_{k,v}}{N_k}} ,
\end{align}
where $N_k$ is the number of instances in the $k$-th subgroup, and $N_{k,v}$ is the number of instances in the $k$-th subgroup with value $v$.
For example, if all the instances of subgroup $k$ have value $v_{3,1}$ (e.g., India), for the feature $f_3$ (e.g., native country), the feature entropy is 0 and $f_3$ is a dominant feature for the subgroup.

\subsubsection{Displaying Suggested Subgroups}

\begin{figure*}
    \centering
    \includegraphics[width=0.93\textwidth,keepaspectratio]{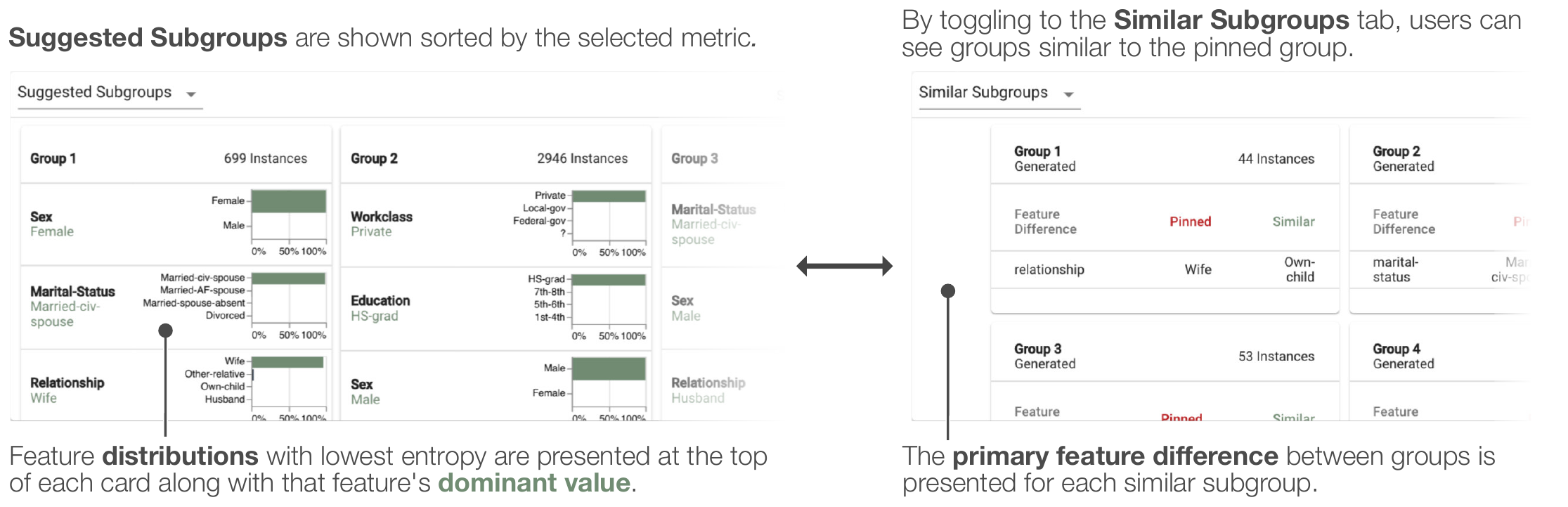}
    \vspace{-12pt}
    \caption{Here we can see the~\suggest~for both suggested and similar subgroups.
    Users can hover over any card to see detailed feature and performance information in the~\detail.
    }
    \label{fig:suggest}
\end{figure*}

We display the generated subgroups in the~\suggest~at the bottom of
the interface, as seen in~\autoref{fig:suggest}. 
Since the generated subgroups are not strictly defined
by a few features, it is important to show the feature distributions
for each feature in a group.  
Each suggested subgroup has a list of its features and
dominant value, along with a histogram of the value distribution for
each feature.  
The features are sorted according to their dominance,
with the dominant value being displayed under the feature name.  
This interface allows users to see what values make up a subgroup and develop an idea of which subgroups may be underperforming.

To explore the groups, users can filter and sort the groups to refine their search space (C3).
Since users may find certain metrics more important than others for certain problems, they can choose which metrics to sort the suggested groups by in ascending order (C5).
For example, if for a given problem recall is an important metric, users can 
find generated subgroups with the lowest recall.

Furthermore, users can use the same size slider used to filter the~\strip~by size to filter the generated subgroups. 
Similar to the reasoning for filtering by size in the strip plot, very small groups may not be large enough to draw statistically significant conclusions from.
Filtering the groups can remove noise and help users further refine their search space of problematic groups.

Users can hover over a suggested subgroup card to show its detailed performance metrics in the~\detail~and add the group to the~\strip.
If a user wants to investigate the group further, they can click on the card, pinning the group and allowing them to compare it to other groups or export it for sharing. 

\subsection{Similar Subgroups [G4]} \label{sec:similar}
Once a user has discovered an interesting subgroup, it can be helpful to look at similar subgroups to either investigate the impact of certain features or to find more general groups with performance issues (C4).
Finding similar groups is difficult since it is not a well defined task and can require searching a combinatorially large space.

To formalize similarity and refine the subgroup search space, we apply ideas from statistics and machine learning explainability to this task. 
When comparing suggested subgroups, we use similarity in the form of statistical divergence to compare how closely related groups are.
For user-specified subgroups, we apply the concept of counterfactual explanations by finding groups with minimal value differences that have significantly different performance.

\subsubsection{Finding Similar Subgroups}

Similarity between subgroups can be thought of as the statistical distance between the feature distributions of groups; the more values two subgroups share, the more similar we consider them.
Statistical distance can be measured in a variety of ways, but we found Jensen-Shannon (JS) divergence to be a good measure for our use case.
As a derived form of Kullback-Leibler divergence, JS divergence is a similar measure with the benefits of being bi-directional and always having a finite value.
Since we often have zero-probability values, JS divergence makes calculating statistical similarity more straightforward and standardized.

We calculate similarity between groups by summing the JS divergence between all features for a pair of subgroups.
This sum gives us a measure of how similar two subgroups are on aggregate.
Formally, we calculate the total distance $D$ between subgroups $k$ and $k'$ as follows, where $G_{k, f}$ represents the value distribution of feature $f$ in subgroup  $k$:
\vspace{-7pt}
\begin{align}
    D (k, k') = \sum_{f \in \mathcal{F}} \text{JS} (G_{k,f} || G_{k',f}) .
\end{align} 
This definition of subgroup similarity applies most directly to the suggested subgroups that have some distribution over values for each feature.
When comparing two suggested subgroups against each other, we can use the formal definition of JS divergence and sum the average distance of their feature distributions.
For comparing user-specified and suggested subgroups against each other we can use a similar technique with a small optimization --- since user-specified subgroups will have 0 probability for all values but the selected values in each feature, it is only necessary to calculate the JS divergence for the values present in the user-specified group.

\textbf{User-specified subgroup comparison.} 
The final potential case for comparison is between two user-specified subgroups.
The use of JS divergence as a measure of similarity begins to break down and lose its utility for this use case.
The divergence will only ever be one when groups have the same value for a feature or zero when they do not.
This metric in practice just counts the number of features with the same value between two groups.
While this measure provides some information about subgroup similarity, it is not as informative or accurate as it is when comparing distributions over features in the other two cases.

To provide a more useful comparison of groups, we use the idea of counterfactual explanations~\cite{wachter2017counterfactual}
which are usually presented in the following form: What are the minimum number of features we have to change to switch the classification of an instance?

Since we are looking at subgroups of multiple instances instead of individual examples, we use a modified notion of counterfactuals for comparing user-specified subgroups:
If we only switch one or two feature values for a subgroup, which similar groups have the most surprising changes in performance?
This question can help users answer similar questions as they would for the groups found using JS divergence.

\subsubsection{Displaying Similar Subgroups}

Once similar subgroups have been found for a selected subgroup, we reuse the~\suggest~from \autoref{sec:suggest} to display the groups to the user.
Each subgroup is represented by a card containing a group number and the size of the subgroup.
Since selecting a subgroup displays its information in the~\detail, only the information most pertinent to deciding which subgroup to investigate should be displayed.

Continuing with the philosophy of treating similar groups as counterfactuals, we display the primary feature difference between two groups in the case of user-specified subgroups, and the most divergent feature for suggested subgroups.
By displaying the feature difference, we emphasize the importance of that feature in the performance difference between the groups.

The same two primary interactions are available for exploring similar groups: 
sorting and filtering (C3).
Users can sort the groups by any fairness metric and filter the groups by size.
As with the strip plot and suggested views, this mechanism helps users find statistically significant subgroups that the model is underperforming for in metrics the user finds important.

\textbf{Similar subgroup importance.} Similar subgroups can be informative in two primary manners: finding features which are important for performance and discovering more general subgroups.
Given that we are looking at two similar subgroups, they likely only differ in one or two features.
If the performance between these two groups is vastly different, it is indicative that the features which are different may contribute significantly to performance (C6).
On the other hand, if the two groups have very similar performance, it may mean that a broader subgroup not split using the differing features is also underperforming and should be analyzed.

\subsection{Detailed Subgroup Analysis and Comparison [G5]} \label{sec:detail}

The final step in discovering and formalizing group inequity is to examine the details of a subgroup's features and performance.
We enable this interaction with the~\detail~on the right hand side of the system.

A user is able to see the details for two groups in the~\detail, the pinned and hovered group.
A group can be pinned when a user clicks on it in the~\strip~or~\suggest, and is designated by a light red across the UI.
The hovered group is designated by a light blue across the UI.
These two distinct colors allow users to see a selected group's information across various different views.

\begin{figure}[t]
    \centering
    \includegraphics[width=\linewidth,height=8.0cm,keepaspectratio,trim={0 .74in 0 0},clip]{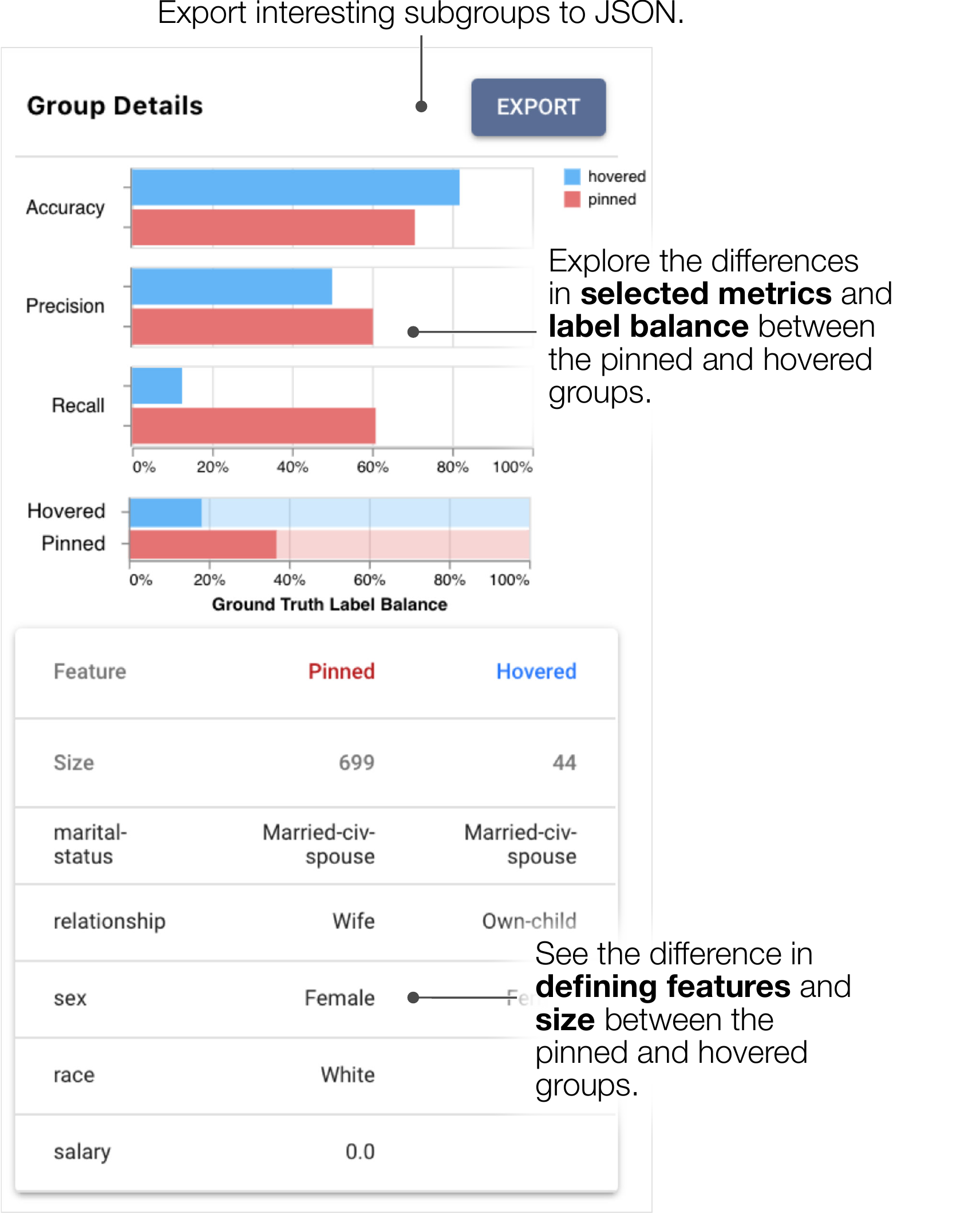}
    \vspace{-10pt}
    \caption{In the~\detail~users can compare the performance and makeup of the pinned and hovered subgroups, providing insight into the causes of performance differences. }
    \label{fig:subgroup_comparison}
\end{figure}

There are three primary components in the~\detail, as seen in~\autoref{fig:subgroup_comparison}.
The topmost component is a bar chart displaying how a group performs for selected performance metrics.
While users can see the values of the fairness metrics in the strip plot, the bar chart allows users to see the specific values and enables comparison between groups with a grouped bar chart (C5).
The grouped bar chart also enables direct comparison between the pinned and hovered subgroups without the distraction of other groups.

The second component in the~\detail~is a bar chart for the ground truth label balance of both selected subgroups. 
The label imbalance is important because it can often explain extreme values for metrics like recall and precision and can suggest reasons for bias (C6).
For example, a subgroup with 95\% negative values can get a 95\% accuracy by classifying everything as negative, even though it will have a 0\% sensitivity.

The final subgroup comparison interface is a table delineating and comparing the features of the pinned and hovered subgroups.
For user-specified subgroups, this table shows the features and values that define the subgroup.
For suggested subgroups, this shows the top 5 dominant feature values for that group, and users can see the full distribution in the~\suggest~view.

\begin{figure}[t]
    \centering
    \includegraphics[width=0.6\linewidth,trim={0 .02in 0 .19in},clip]{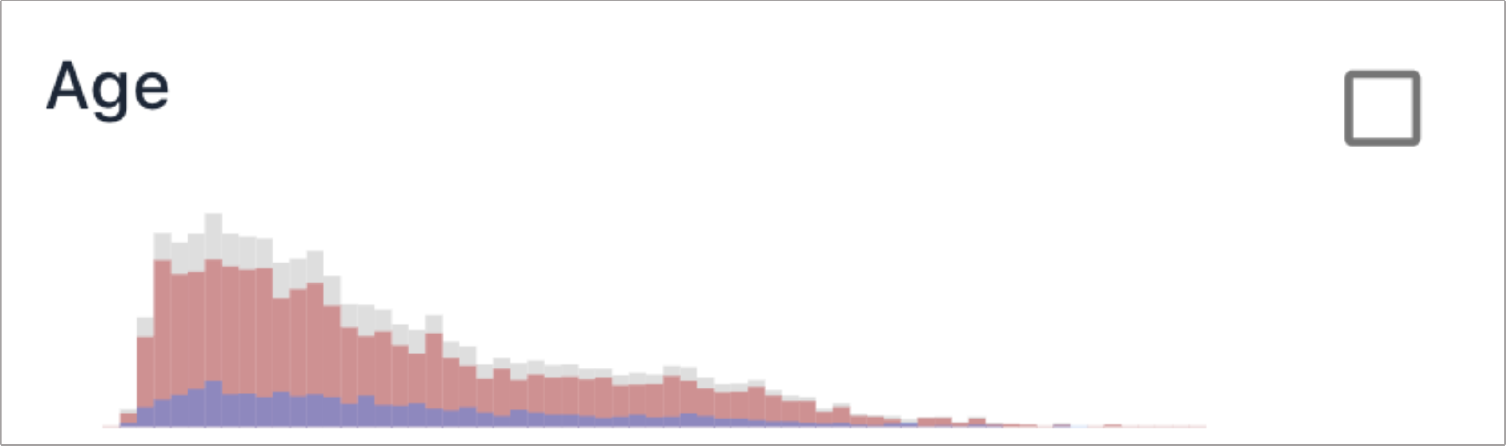}
    \vspace{-11pt}
    \caption{When groups are \textbf{\textcolor{sysred}{pinned}} and \textbf{\textcolor{sysblue}{hovered}}, users can compare their feature distributions in the~\feature~.
    }
    \label{fig:distribution}
\end{figure}

\textbf{Subgroup feature distributions.} There is additional information about the pinned and hovered subgroup in the~\feature. 
When a subgroup is hovered or pinned, a histogram of each feature's distribution for that group is overlaid on the overall distribution (C2).
When there is both a pinned and hovered subgroup, the histograms are overlaid with opacity, allowing users to see how similar the distributions are (\autoref{fig:distribution}).

The distribution of a subgroup's features can be an important indicator of why a subgroup is underperforming and suggest potential resolutions (C6).
If a subgroup's ground truth labels are well balanced, there should be some diversity in the other features of a subgroup for the classifier to be able to discriminate between the two labels.
For example, if all White males are also high school educated, married, and from the United States, and they are split between positive and negative classes, it is nearly impossible for a classifier to accurately predict the class for anyone in that subgroup.

An extra interaction in the~\detail~is an export button for sharing a discovered subgroup.
Once a user has found subgroups of interest, they can export the pinned and hovered subgroups to a JSON file with their composition and metrics.

\section{Use Cases}\label{use-cases}
In this section, we describe how \system{} can be used in practice to audit models after they have been trained with two example usage scenarios. 
The first scenario highlights how \system{} can be used to audit models for biases against known vulnerable groups in the context of a recidivism prediction system.
The second use case shows how users without previous knowledge or intuitions about potential biases can use the system to find issues, for this example with an income prediction model. Both of these use cases utilize real world datasets to demonstrate the applications of our system.

\subsection{Auditing for Known Biases in Recidivism Prediction}\label{case-1}
\newcommand{\ds}{data\xspace scientist\xspace}

For our first example use case, we will demonstrate how \system could be used to discover biases in a classifier for recidivism prediction used in the context of deciding who should be given bail.
In this use case, we use a classifier based on data gathered by ProPublica about the real-world tool, COMPAS, that assigns risk scores to criminals to determine their likelihood of re-offending~\cite{propublica_2019}. 
The original dataset ranks risk from 1-10, with risks from 1-4 constituting "low" risk, those from 5-7 constituting "medium" risk, and those from 8-10 as "high" risk.  
Following the same methodology as in the ProPublica analysis, we formulate this as a binary classification task by taking risk scores above "low" (i.e. above 4) as positive model predictions to re-offend, and those at 4 or below as negative predictions as any prediction of risk above low indicates COMPAS is predicting recidivism.
Ground-truth labels correspond to whether a defendant released on bail was arrested for another crime within 2 years of their release. 
An audit by ProPublica revealed that the COMPAS tool is biased to give higher risk scores and thus predict a higher rate of recidivism for Black defendants than other races \cite{angwin2016bias}. 
Here, we will demonstrate how a data scientist auditing their model in \system could arrive at the same conclusion.

\textbf{Known subgroup auditing.} To begin their audit, a data scientist would load the COMPAS dataset along with model predictions and ground truth labels into \system. 
Given their domain knowledge, the \ds is aware that, in previous applications involving recidivism prediction, many tools have displayed imbalanced performance for certain genders and races. 

To test whether differing performance holds for this model and dataset, the \ds uses
the~\feature~to generate all intersectional subgroups of race and sex. 
When the groups are added to the~\strip~ (\autoref{fig:teaser}B), she immediately sees that the groups are spread out broadly across various metrics, suggesting this model may have very different predictive performance on different subgroups. For instance, as we can see in \autoref{fig:teaser}B (top row), the different intersectional subgroups of sex and race have accuracies ranging from around 50\% to 100\%.

While the \ds is interested in the accuracy of her model, she cares most about whether her model has large intra-group variation in terms of its false positive rate. 
For this model, this translates to how many of the people who are not risky are classified as risky. Additionally, she wants to know if these mistakes are distributed unevenly across the different demographic groups.
A high false positive rate for this model indicates that many low-risk 
people (who might be good candidates for release on bail) would be labeled as high-risk by the model. 
If this model were used to help determine whether a person was seriously considered for release, false positives would correspond to low-risk candidates for release who might be passed over for bail.

To audit the false positive performance metric, the \ds adds a strip plot for it using the metric selector shown at the top of \autoref{fig:teaser}B. 
She then hovers over the bar in the false positive rate strip plot (the bottom row in \autoref{fig:teaser}B) with the highest value, and sees that this corresponds to the African-American males subgroup with a 43\% false positive rate (colored in \textcolor{sysblue}{blue}) compared to the dataset average of around 29\%. 
The \ds pins this subgroup by clicking on this group's strip in the~\strip~to investigate it further and compare it to other groups. 

By hovering over the other subgroups, she can compare the base rate of recidivism for the pinned group of African-American males relative to other groups. Looking at the Ground Truth Label Balance in \autoref{fig:teaser}C, we see that the base rate for African-American Males (blue) is almost 60\% positive (i.e. 60\% rate of recidivism in ground truth), whereas for Caucasian males (red) it is just over 40\%.

Thus, if a model makes only one prediction for the entire subgroup of African-American Males, choosing to label the subgroup as positive (a prediction of high recidivism risk) will have higher accuracy than for other subgroups. 
Less extreme versions of this statement may still hold: to maximize accuracy for this subgroup, a model will use a larger number of positive labels than negative labels. Since the \ds has noticed that the African-American male subgroup has a very high base rate, but also the highest False Positive Rate out of any of the subgroups in view and still has an accuracy very similar to that of Caucasian Males, she thinks this part of her model needs to be altered to give more equitable results.

Here, our example data scientist had suspicions about groups the model might be biased against and was able to leverage \system to empirically confirm these suspicions. 
From here, she could use the export function in the system to save these subgroups and devise a plan for corrective action for this model or dataset. 

\textbf{Investigating Suggested Subgroups.} Although our data scientist was able to use her domain knowledge to inform her subgroup selection at first, she is interested in whether the model also contains biases against other intersectional subgroups. 
To aid in the exploration, this data scientist would turn to the~\suggest~panel to find other potentially problematic groups. 

\begin{figure}
    \centering
    \includegraphics[width=0.98\linewidth,trim={0 .35in 0 0},clip]{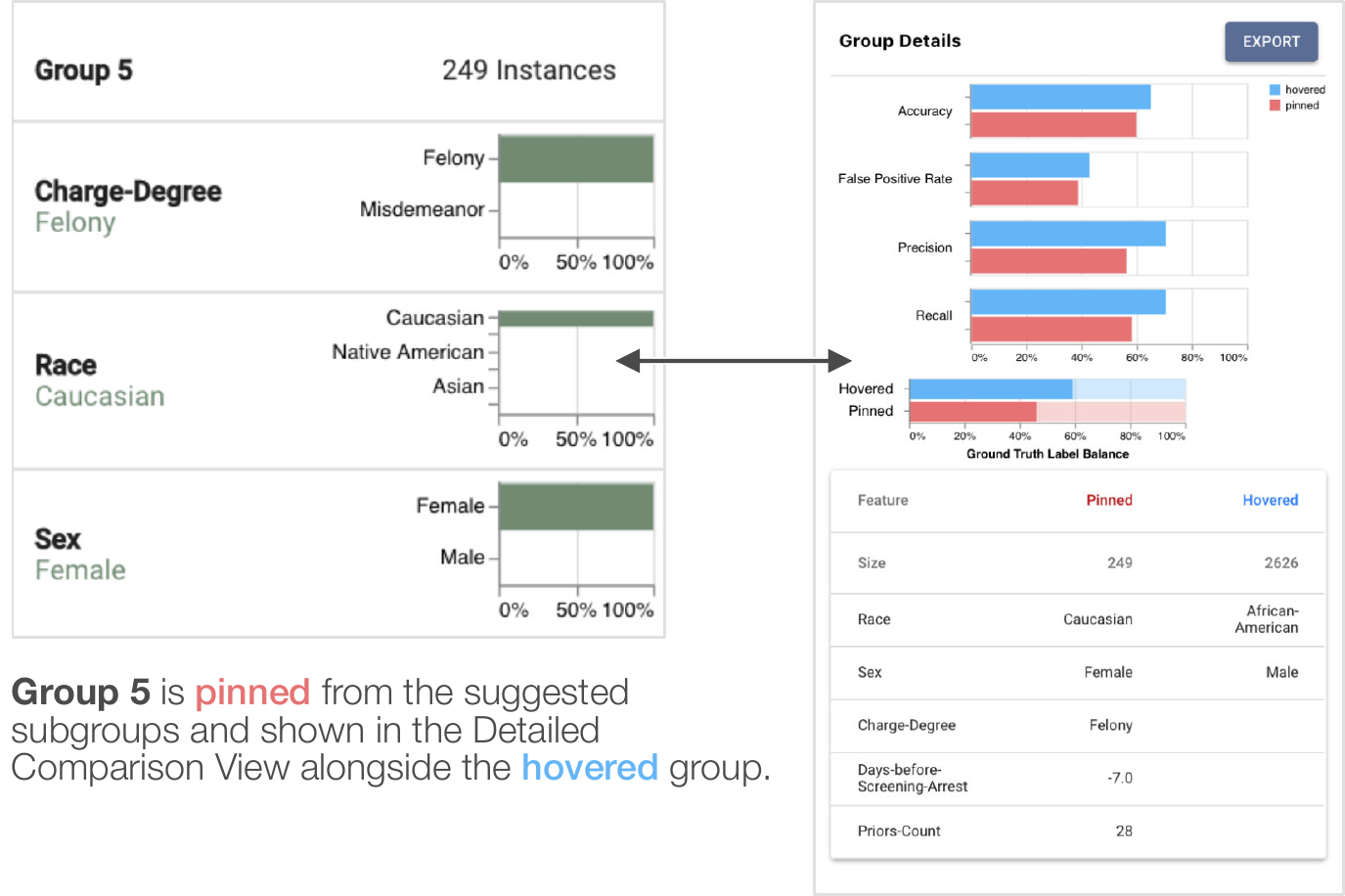}
    \vspace{-11pt}
    \caption{A user investigates an interesting subgroup discovered in the~\suggest.}
    \label{fig:use-1-suggested}
\end{figure}

The \ds first sorts the suggested groups by their false positive rate, since she is most worried about that metric.
While the first few groups with the highest false positive rate are made up of African-American males, corroborating her earlier findings, one of the following groups provides a different result. 

The fifth generated group (\autoref{fig:use-1-suggested}) is relatively large with 249 instances, and has a high false positive rate of 39\%.
By inspecting the composition of this group in the~\detail and the subgroup card, she sees that the most defining characteristics of this group are Caucasian females with a felony charge.
The label imbalance for this group is about 45\% positive and 55\% negative and therefore not as pronounced as the base rate imbalance for African-American males (\autoref{fig:teaser}C).
This gives the \ds two potential hypotheses about sources of this high false positive rate. 
Her first hypothesis is that the rather small group was not large enough to have been given priority in training; the second is that the class of models considered during training may have been too simple to express the difference between classes in this subgroup.
These observations allow our data scientist to make more informed decisions in how to best change her model to address these disparities. 

\subsection{Discovering Biases in Income Prediction}\label{case-2}

Next, let us consider a model used to offer loan forgiveness to individuals based off their annual income. 
Our \ds in this situation does not have access to people's annual income so hopes to use demographic information to predict income.
She therefore trains a model on the UCI Adult Dataset~\cite{Dua:2017} to predict whether or not someone makes under \$50,000 a year, allowing her to allocate loan forgiveness to lower income candidates with higher fidelity.

\textbf{Model training.} 
After testing different types of models and hyperparameters, our \ds finds that a two-layer neural network performs best, with an overall accuracy of 85\%.
While encouraged by the high accuracy of her model, the \ds is aware of recent news of algorithmic bias and wants to ensure that her model is treating different demographic groups with similar predictive performance.
She decides to audit her model using \system, and loads her dataset, labels, and model predictions into the system.

\textbf{Dataset exploration and subgroup creation.} When first opening \system, the \ds uses the~\feature~on the left to look at how balanced her dataset is. 
While she is unaware of any biases in her data, she immediately notices from looking at the feature histograms that the dataset has a disproportionate representation of males, with males making up more than 2/3 of all instances (see \autoref{fig:drawer}).
To investigate the impact of this imbalance, she selects the feature for sex to generate male and female subgroups.
When looking at these two subgroups, she sees in the~\strip~that there is a gap of almost 10\% in model accuracies between the male and female subgroups (top of \autoref{fig:stripfig}). Despite the higher accuracy of the female subgroup, she notices that the male subgroup has a higher value for precision and recall.

\textbf{Suggested subgroups.}
After seeing the fairly large gap in the accuracy of her model between subgroups defined by just one feature, the \ds is curious about what other combinations of features might lead to poor performance in her model.
She turns to the~\suggest~to see what she can find.
Keeping the default sorting of groups by lowest accuracy, she notices that suggested Group 1 (shown on the left side of \autoref{fig:suggest}) has an accuracy of around 71\%, far below the dataset average of 85\%. 
By inspecting the feature distribution charts in the~\suggest, she sees that this group is primarily defined by Females with a marital status of ``Married-civ-spouse" and relationship status of ``Wife" as shown by the value distribution graphs in Group 1 of \autoref{fig:suggest}.
Since she wants to better understand why her model is performing poorly for this group, the \ds tries exploring similar groups.

\textbf{Similar subgroups.}
Using her discovery from the Suggested Subgroups tab, our \ds wants to see how groups of females compare to one another across the ``marital-status'' and ``relationship'' features. 
She generates these subgroups in the~\feature~and pins suggested subgroup 1 from earlier to inspect similar groups. 
Here, she notices that the similar group with the  lowest accuracy is the one comprised of females with a marital status of ``married-civ-spouse'' but a relationship of ``own-child''. This group is quite small with only 44 instances. 

To see how this group fits into the overall dataset, the \ds looks to the~\feature. Here, she sees that ``married-civ-spouse'' is the most common value for the Marital-Status feature, and ``own-child'' is the third most common value for the Relationship feature. These features combine to make a subgroup with relatively few values in the dataset. 

When looking at the~\detail~for this similar subgroup, the \ds notices that the base rate for the ``Female, own-child, married-civ-spouse'' subgroup is heavily skewed to less than 20\% positive ground truth instances (\autoref{fig:subgroup_comparison}). The \ds therefore hypothesizes that the low accuracy for this group may be due to its small size and the skewed base rate.
The \ds notes these observations and aims to gather more data and try using a more expressive model to see if she can address these discrepancies.

\section{Technical Implementation}

\system{} is a web-based system built using the open-sourced JavaScript framework \textit{React}. 
Many additional libraries were used for building the system, including \textit{D3.js} and \textit{Vega Lite} for visualizations and \textit{Material.ui} for visual components and interface style. Scripts for pre-processing and clustering were written in Python and use the \textit{scikit-learn} implementation of K-means.

\section{Limitations and Future Work}

\quad \textbf{Improving and measuring the effectiveness of the subgroup generation technique.} 
While we found that the generated subgroups often provide useful suggestions, we hope to test whether these generated groups align well with groups users find important in future work.
Collecting labeled data of datasets with outputs and important underperforming subgroups would allow us to quantify the effectiveness of our technique. 
Additionally, we plan to experiment with more clustering techniques, such as subspace clustering methods \cite{ParsonsSubspace} to future versions of \system{} so that users can see how the groups compare. Especially in high dimensional data, subspace clustering has the potential to reveal interesting groups with poor performance that are primarily defined by only a few features.

\textbf{Supporting more types of problems and data.} \system{} currently only supports binary classification and tabular data.
The current interface can be expanded to support multiclass classification, but additional visualizations views would need to be added for regression.
It would additionally be nice to support some sort of graphical or textual data.
The current interface works if the outputs of image classification are loaded with demographic data, but enabling the display of images could aid in auditing groups.

\textbf{Scaling to millions of instances.} The current implementation of \system{} is able to scale to tens and hundreds of thousands of data points, but does not support even larger datasets very well.
We are looking at improving the efficiency of the subgroup generation and suggestion technique to enable our system to continue to work in browser while at scale.

\textbf{Suggesting and providing automatic resolutions.} Various techniques exist to address bias in machine learning, many of which can be applied as a post-processing step to the output of a classifier.
In addition, there are patterns as to what the potential reasons for bias are which could be learned by a model or codified into heuristics.
We aim to implement some of the post-processing steps into \system{} and add capability to highlight and suggest potential issues.

\acknowledgments{
This work was supported by NSF grants IIS-1563816, CNS-1704701, and TWC-1526254, a NASA Space Technology Research Fellowship, and a Google PhD Fellowship.
}

\bibliographystyle{abbrv-doi-hyperref-narrow}

\bibliography{refs}
\end{document}